\def\year{2020}\relax
\documentclass[letterpaper]{article} % DO NOT CHANGE THIS
\usepackage{aaai20}  % DO NOT CHANGE THIS
\usepackage{times}  % DO NOT CHANGE THIS
\usepackage{helvet} % DO NOT CHANGE THIS
\usepackage{courier}  % DO NOT CHANGE THIS
\usepackage[hyphens]{url}  % DO NOT CHANGE THIS
\usepackage{graphicx} % DO NOT CHANGE THIS
\usepackage{graphicx}  % DO NOT CHANGE THIS
\usepackage{algorithm}
\usepackage{multirow}

\title{Consistent and Complementary Graph Regularized\\ Multi-view Subspace Clustering}
\author{Qinghai Zheng,\textsuperscript{\rm 1} Jihua Zhu,\textsuperscript{\rm 1}\thanks{Corresponding Author} Zhongyu Li,\textsuperscript{\rm 1} Shanmin Pang,\textsuperscript{\rm 1}  Jun Wang,\textsuperscript{\rm 2} Lei Chen\textsuperscript{\rm 3}\\  % All authors must be in the same font size and format. Use \Large and \textbf to achieve this result when breaking a line
\textsuperscript{\rm 1}School of Software Engineering, Xi'an Jiaotong University, Xi'an 710049, China\\
\textsuperscript{\rm 2}Shanghai Institute for Advanced Communication and Data Science,\\ School of Communication and Information Engineering,Shanghai University, Shanghai 200444, China\\
\textsuperscript{\rm 3}Jiangsu Key Laboratory of Big Data Security and Intelligent Processing,\\Nanjing University of Posts and Telecommunications, Nanjing 210023, China}
\begin{document}

\maketitle

\begin{abstract}
This study investigates the problem of multi-view clustering, where multiple views contain consistent information and each view also includes complementary information. Exploration of all information is crucial for good multi-view clustering. However, most traditional methods blindly or crudely combine multiple views for clustering and are unable to fully exploit the valuable information. Therefore, we propose a method that involves consistent and complementary graph-regularized multi-view subspace clustering (GRMSC), which simultaneously integrates a consistent graph regularizer with a complementary graph regularizer into the objective function. In particular, the consistent graph regularizer learns the intrinsic affinity relationship of data points shared by all views. The complementary graph regularizer investigates the specific information of multiple views. It is noteworthy that the consistent and complementary regularizers are formulated by two different graphs constructed from the first-order proximity and second-order proximity of multiple views, respectively. The objective function is optimized by the augmented Lagrangian multiplier method in order to achieve multi-view clustering. Extensive experiments on six benchmark datasets serve to validate the effectiveness of the proposed method over other state-of-the-art multi-view clustering methods.
\end{abstract}

\section{Introduction}
Clustering is an important task in unsupervised learning, which can be a preprocessing step to assist other learning tasks or a stand-alone exploratory tool to uncover underlying information from data \cite{zhou2012ensemble_methods}. The goal of clustering is to group unlabeled data points into corresponding categories according to their intrinsic similarities. Many effective clustering algorithms have been proposed, such as k-means clustering \cite{k_means}, spectral clustering \cite{spectral_clustering_tutorial} and subspace clustering \cite{subspace_clustering,sparse_subspace_clustering_PAMI2013}. However, these methods are designed for single-view rather than multi-view data from various fields or different measurements common in many real-world applications. Unlike single-view data, multi-view data contains both the consensus information and complementary information for multi-view learning. \cite{multiview_learning_survey}. Therefore, an important issue of multi-view clustering is how to fuse multiple views properly to mine the underlying information effectively. Evidently, it is not a good choice to use a single-view clustering algorithm on multi-view data straightforward \cite{co_reg_spectral,multiview_learning_survey,multiview_clustering_survey}. In this study, we consider the multi-view clustering problem based on the subspace clustering algorithm \cite{subspace_clustering,LRR_PAMI2012}, owing to its good interpretability and promising performance in practice.

Multi-view subspace clustering assumes that all views are constructed based on a shared latent subspace and pursues a common subspace representation for clustering \cite{multiview_clustering_survey}. Many multi-view subspace clustering methods have been proposed in recent years \cite{DiMSC,CSMSC_AAAI2018,multiview_subspace_clustering_TangChang_TMM2018,MLRSSC_PR2018,GLMSC_PAMI2018,multiview_subspace_clustering_dual_ZhouTao}. Although good clustering results can be obtained in practice, there are some deficiencies in the existing methods. First, some methods deal with multiple views separately and combine clustering results of different views directly. As a result, the relationship among multiple views is ignored during the clustering process. Second, most existing methods only take the consensus information or the complementary information of multi-view data into consideration rather than explore both of them. Third, a few methods integrate graph information of multiple views into the subspace representation for improving clustering results, however, only the first-order similarity \cite{second_order_proximity,first_order_prixitmity} of data points in multi-view data is considered and employed as is, which is oversimplified for multi-view clustering. Actually, the first-order similarity is an observed pairwise proximity, with the local graph information lacking in the global graph structure \cite{second_order_proximity}. Moreover, the clustering structure of the first-order proximity has often discordance among different views, because different views have different statistic properties.

\begin{figure*}[t]
\centering
\includegraphics[width=0.9\textwidth]{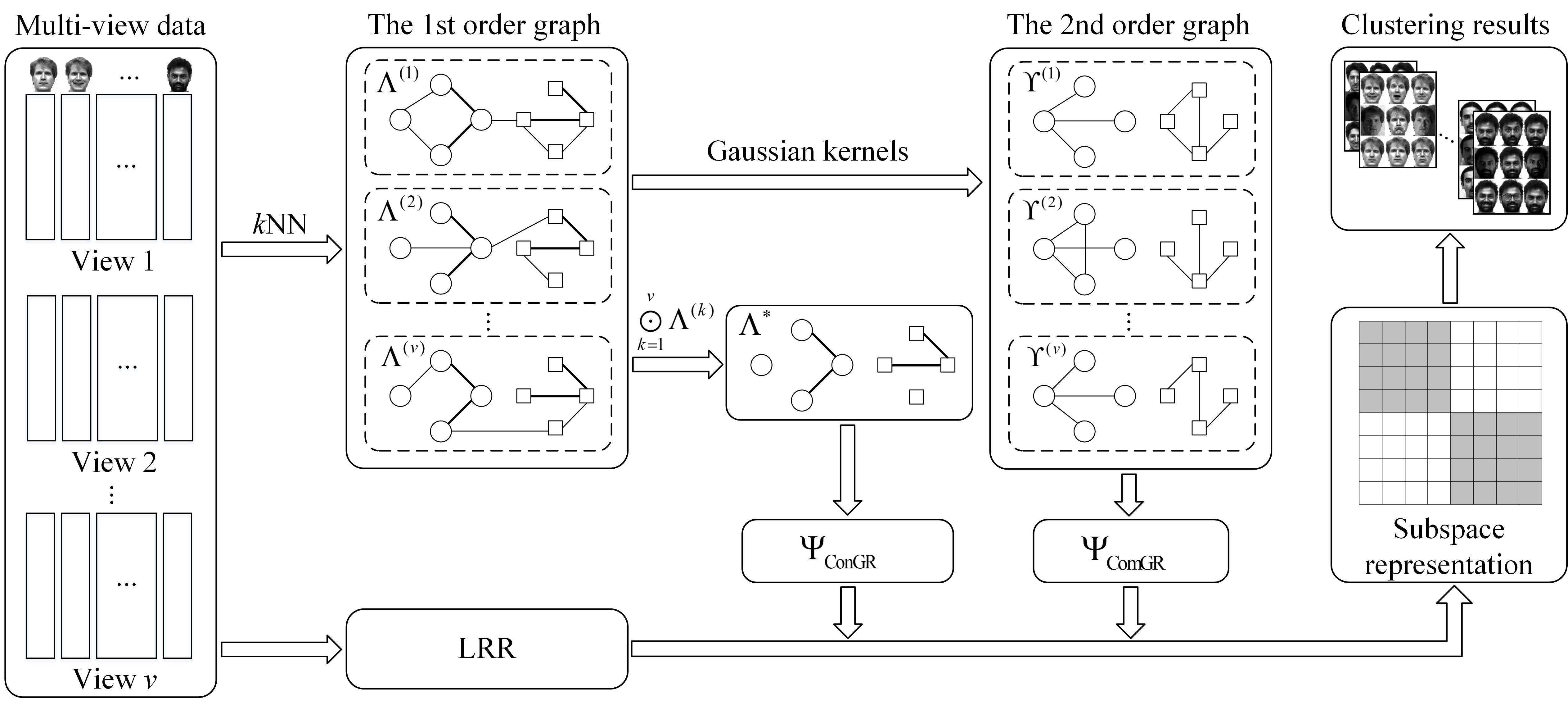} % Reduce the figure size so that it is slightly narrower than the column.
\caption{Illustration of the proposed method. $\left\{ {{{\bf\Lambda} ^{(k)}}} \right\}_{k = 1}^v$ and $\left\{ {{{\bf\Upsilon} ^{(k)}}} \right\}_{k = 1}^v$ are the first order proximity and second order proximity. All views are processed simultaneously, and graph information is also considered for multi-view clustering by introducing the consistent regularizer ${\Psi _{{\rm{ConGR}}}}$ and the complementary regularizer ${\Psi _{{\rm{ComGR}}}}$.}
\label{Framework_GRMSC}
\end{figure*}

To address the above-mentioned limitations of the existing clustering methods, a graph-regularized multi-view subspace clustering (GRMSC) methods is presented in this study. Considering that clustering results should be unified across different views, it is vital for multi-view clustering to integrate information of multiple views in a suitable way \cite{multiview_learning_survey,multiview_clustering_survey}. In the proposed method, low-rank representation (LRR) \cite{LRR_PAMI2012} is performed on all views jointly, and a common subspace representation is obtained and accompanied with two graph regularizers: a consistent graph regularizer based on the first-order proximity to explore the consensus information of all views, and a complementary graph regularizer based on the second-order proximity to explore the complementarity of different views. Figure \ref{Framework_GRMSC} illustrates the complete framework for the proposed method. The consistent and complementary graph regularizers are discussed in detail consequently. To achieve multi-view clustering, an algorithm based on the augmented Lagrangian multiplier (ALM) method \cite{ALM} is designed to optimize the proposed objective function. Finally, clustering results are achieved by applying spectral clustering on the affinity matrix calculated based on the common subspace representation. Comprehensive experiments on six benchmark datasets are conducted to validate the superior performance of the proposed multi-view clustering method compared with the existing state-of-the-art clustering methods. 

The main contributions of this study are as follows:
\begin{itemize}
	\item [1)] 
	A novel GRMSC method is proposed to perform clustering on multiple views simultaneously by fully exploring the intrinsic information of multi-view data;
	\item [2)] 
	A consistent graph regularizer and a complementary graph regularizer are introduced to integrate the multi-view information in a suitable way for multi-view clustering; 
	\item [3)] 
	An effective algorithm based on the ALM method is developed and extensive experiments are conducted on six real-world datasets to confirm the superiority of the proposed method.
\end{itemize}

\section{Related Works}
In recent years, many multi-view clustering methods have been proposed. Based the way the views are combined, most existing methods can be classified roughly into three groups \cite{multiview_learning_survey}: co-training or co-regularized, graph-based, and subspace-learning-based methods.

Multi-view clustering methods of the first type \cite{co_reg_spectral,co_training_spectral,zhailin_2019multi} often combine multiple views under the assumption that all views share the same common eigenvector matrix \cite{multiview_learning_survey,multiview_clustering_survey}. For example, co-regularized multi-view spectral clustering \cite{co_reg_spectral} learns the graph Laplacian eigenvectors of each view separately, and then utilizes them to constrain other views to obtain the same clustering results. The graph-based method \cite{RMSC,AMGL,GSF,MCGC,GMC} explores the underlying information of multi-view data by fusing different graphs. For instance, robust multi-view subspace clustering (RMSC) \cite{RMSC} pursues a latent transition probability matrix of all views via low rank and sparse decomposition, and then obtains clustering results based on the standard Markov chain. Auto-weighted multiple graph learning (AMGL) \cite{AMGL} integrates all graphs, with auto-weighted factors based on the fact that different views are associated with incomplete information for real manifold learning and have the same clustering results. Multi-view consensus graph clustering (MCGC) \cite{MCGC} achieves clustering results by learning a common shared graph of all views with a constrained Laplacian rank constraint. Graph-based multi-view clustering (GMC) \cite{GMC} introduces an auto-weighted strategy and a constrained Laplacian rank constraint to construct a unified graph matrix for multiple views. Many multi-view subspace clustering approaches \cite{DiMSC,LtMSC,MVSC,LMSC,tSVDMSC,MLRSSC_PR2018} have been proposed as well based on the idea that multiple views have the same latent subspace and a common shared subspace representation. Low-rank tensor-constrained multi-view subspace clustering (LT-MSC) \cite{LtMSC} and tensor-singular value decomposition based multi-view subspace clustering (t-SVD-MSC) \cite{tSVDMSC} seeks the low-rank tensor subspace to explore the high-order correlations of multi-view data for clustering fully. Latent multi-view subspace clustering (LMSC) \cite{LMSC} seeks an underlying latent representation, which is the origin of all views, and runs the low-rank representation algorithm on the learning latent representation simultaneously. Multi-view low-rank sparse subspace clustering (MLRSSC) \cite{MLRSSC_PR2018} aims to learn a joint subspace representation and constructs a shared subspace representation with both the low-rank and sparsity constraints.

Even though the various multi-view clustering methods are based on different theories, the key objective of them all is one, i.e., achieving promising clustering results by combining multiple views properly and exploring the underlying clustering structures of multi-view data fully. Unlike most existing methods, the method proposed in this study integrates the first- and second-order graph information into the multi-view subspace clustering process by introducing a consistent graph regularizer and a complementary graph regularizer so that both consensus information and complementary information of multi-view data can be explored simultaneously.

\section{The Proposed Approach}

In this section, we discuss the GRMSC approach. Figure \ref{Framework_GRMSC} presents the complete framework for the proposed method, and Table \ref{table_symbol} presents the symbols used in this paper.

\begin{table}[t]
\caption{Main symbols employed in this paper.}\smallskip
\centering
\resizebox{0.95\columnwidth}{!}{
\smallskip\begin{tabular}{l|l}
\hline
Symbol & Meaning \\
\hline
$n$ & The number of samples. \\
$v$ & The number of views. \\
$c$ & The number of clusters.\\
$d_i$ & The dimension of the $i$-th view.\\
${{\bf X}^{(i)}} \in {R^{{d_i} \times n}}$ & The data matrix of the $i$-th view.\\
${{\bf X}_j^{(i)}} \in {R^{d_i}}$ & The $j$-th data point from the $i$-th view.\\
${{\bf A}_i}$ & The $i$-th column of matrix ${\bf A}$.\\
${\left\| {\bf A} \right\|_{2,1}}$ & The $l_{2,1}$ norm of matrix ${\bf A}$.\\
${\left\| {\bf A} \right\|_ * }$ & The trace norm of matrix ${\bf A}$.\\
${\left\| {\bf A} \right\|_ F^2 }$ & The Frobenius norm of matrix ${\bf A}$.\\
${Tr({\bf A}) }$ & The trace of matrix ${\bf A}$.\\
\hline
\end{tabular}
}
\label{table_symbol}
\end{table}

Given the multi-view data $ {\bf X} = \{ {{\bf X}^{(k)}}\} _{k = 1}^v$, samples of which are drawn from $c$ multiple subspaces, the proposed method can be decomposed into three parts: the low-rank representation on multiple views, consistent graph regularizer, and complementary graph regularizer. The methods can process all views simultaneously, and the intrinsic information can be fully explored.

\subsection{Low-Rank Representation on Multiple Views}
Under the assumption that all views have the same clustering results, LRR \cite{LRR_PAMI2012} is performed on all views and a common shared subspace representation is achieved. Consequently, an optimization problem can be written as follows:
\begin{equation}
\begin{array}{l}
\mathop {\min }\limits_{{\bf Z},{{\bf E}^{(k)}}} {\kern 1pt} {\kern 1pt} {\kern 1pt} {\left\| {\bf Z} \right\|_ * } + {\lambda }\sum\limits_{k = 1}^v {{{\left\| {{{\bf E}^{(k)}}} \right\|}_{2,1}}} \\
{\rm{s}}{\rm{.t}}{\rm{.}}{\kern 1pt} {\kern 1pt} {\kern 1pt} {{\bf X}^{(k)}} = {{\bf X}^{(k)}}{\bf Z} + {{\bf E}^{(k)}},
\end{array}
\label{LRR_MSC_Only}
\end{equation}
where $\bf{Z}$ is the common subspace representation whose columns denote the representation of corresponding samples, ${\bf E}^{(k)}$ indicates the sample-specific error of the $k$th view, and $\lambda$ is the trade-off parameter.

Evidently, the above problem deals with all views simultaneously. However, the information of multiple views cannot be investigated properly in this way, because the low-rank constraint on the common $\bf{Z}$ ignores the specific information of different views. Moreover, the graph information, which is vital for clustering, is not employed in this formulation. A consistent graph regularizer and a complementary regularizer are introduced to handle these limitations.

\subsection{Consistent Graph Regularizer}
Most existing graph-based multi-view clustering approaches employ graphs with first-order proximity for clustering, whose elements denote pairwise similarities between two data points. In this study, Gaussian kernels are utilized to define proximity matrices of all views. Taking the $k$th view as an example, we have the following formula
\begin{equation}
{\bf S}_{ij}^{(k)} = \exp (-\frac{{\left\| {{{\bf X}_i ^{(k)}} - {{\bf X}_j ^{(k)}}} \right\|_2^2}}{{{\sigma ^2}}}),
\end{equation}
where ${\bf S}_{ij}^{(k)}$ denotes the similarity between the $i$th and $j$th data points in the $k$th view, $\sigma $ is the median Euclidean distance. Mutual $k$ nearest neighbor (m-$k$NN) strategy is employed, which means that the elements of the first-order proximity are:
\begin{equation}
{{\bf\Lambda} ^{(k)}} = \left\{ \begin{array}{l}
{\bf S}_{ij}^{(k)},{\kern 1pt} {\kern 1pt} {\kern 1pt} {\rm{if}}{\kern 1pt} {\kern 1pt} {\bf X}_j^{(v)}{\kern 1pt} {\rm{and}}{\kern 1pt} {\kern 1pt} {\kern 1pt} {\kern 1pt} {\bf X}_i^{(v)}{\kern 1pt} {\rm{are}}{\kern 1pt} {\kern 1pt} {\kern 1pt} {\rm{m}\rm{ - }}k{\rm{NN}}{\kern 1pt} {\kern 1pt}, \\
0{\kern 1pt} {\kern 1pt} {\kern 1pt} {\kern 1pt} {\kern 1pt} {\kern 1pt} {\kern 1pt} {\kern 1pt} {\kern 1pt} {\kern 1pt} {\kern 1pt} {\kern 1pt} {\kern 1pt} ,{\kern 1pt} {\kern 1pt} {\rm{otherwise}}
\end{array} \right.
\end{equation}
where ${\bf\Lambda}^{(k)}$ is the first-order proximity matrix of the $k$th view. Clearly, ${\bf\Lambda}^{(k)}$ captures the local graph structures.

However, as shown in Figure \ref{Framework_GRMSC}, the graphs with the first-order proximity among views are different from each other because statistic properties of different views are diverse. Evidently, it is not a suitable way to leverage first proximity matrices straightforward. To explore the common shared intrinsic graph information of multi-view data, a consistent graph regularizer is introduced. Given $ \{ {{\bf\Lambda}^{(k)}}\} _{k = 1}^v$, a proximity matrix ${\bf\Lambda}^*$ can be constructed as follows:
\begin{equation}
{\bf\Lambda}^ * {\rm{ = }}\mathop  \odot \limits_{k = 1}^v {\bf\Lambda}^{(k)},
\end{equation}
where $\odot$ denotes the Hadamard product. It is noteworthy that not all elements of ${\bf\Lambda}^ *$ are taken into consideration. As shown in Figure \ref{Framework_GRMSC}, nonzero elements of ${\bf\Lambda}^ *$ indicate the shared intrinsic consensus graph information of multi-view data. The consistent graph regularizer, i.e., ${\Psi _{{\rm{ConGR}}}}$, for multi-view clustering can be defined as follows:
\begin{equation}
{\Psi _{{\rm{ConGR}}}}(Z) = {\kern 1pt} {\kern 1pt} \frac{1}{2}\sum\limits_{(i,j) \in \Omega } {{\bf\Lambda}_{ij}^ * \left\| {{{\bf Z}_i} - {{\bf Z}_j}} \right\|_2^2},
\label{Consistent_Graph_Regularizer}
\end{equation}
where $\Omega$ is the index set of the nonzero elements in ${\bf\Lambda}^ *$, and we also denote $\bar \Omega$ as the index set of the zero elements in ${\bf\Lambda}^ *$ in future.

The consistent graph regularizer integrates the consensus graph information into the subspace representation properly. For the rest of the parts in graphs of multiple views, a complementary graph regularizer is introduced to explore the complementary information of multi-view.

\subsection{Complementary Graph Regularizer}
Elements in $\bar \Omega$ of $ \{ {{\bf\Lambda}^{(k)}}\} _{k = 1}^v$ are inconsistent across different views. Therefore, it is inadvisable to use them as Eq. (\ref{Consistent_Graph_Regularizer}). How to fuse them effectively is vital for multi-view clustering. In this paper, the second-order proximity matrices of multiple views, i.e., $ \{ {{\bf\Upsilon}^{(k)}}\} _{k = 1}^v$, are introduced, and a complementary graph regularizer is defined to benefit the clustering performance based on the elements in $\bar \Omega$ of $ \{ {{\bf\Upsilon}^{(k)}}\} _{k = 1}^v$.

Under the intuition that data points with more shared neighbors are more likely to be similar, the second-order proximity can be constructed as follows:
\begin{equation}
{{\bf\Upsilon}}_{ij}^{(k)} = \exp (-\frac{{\left\| {{\bf\Lambda}_i^{(k)} - {\bf\Lambda}_j^{(k)}} \right\|_2^2}}{{{\sigma ^2}}}),
\end{equation}
where ${\bf\Upsilon}_{ij}^{(k)}$ denotes the second-order proximity matrix of the $i$th and $j$th data points in the $k$th view. Evidently, the second-order proximity matrices of multiple views, i.e., $ \{ {{\bf\Upsilon}^{(k)}}\} _{k = 1}^v$, capture the global graph information of multi-view data. Furthermore, to investigate the complementary information of multi-view data, the following complementary graph regularizer, i.e., ${\Psi _{{\rm{ComGR}}}}$, is introduced:
\begin{equation}
{\Psi _{{\rm{ComGR}}}}({\bf Z}) = {\kern 1pt} {\kern 1pt} \frac{1}{2}\sum\limits_{k = 1}^v {\sum\limits_{(i,j) \in \bar \Omega } {{\bf\Upsilon}_{ij}^{(k)}\left\| {{{\bf Z}_i} - {{\bf Z}_j}} \right\|_2^2} },
\label{Specific_Graph_Regularizer}
\end{equation}
in which elements in $\bar \Omega$ of $ \{ {{\bf\Upsilon}^{(k)}}\} _{k = 1}^v$ are utilized. Different from the consistent graph regularizer, the complementary graph regularizer defined in Eq. (\ref{Specific_Graph_Regularizer}) explores the global graph information of all views and integrates the complementary graph information into the subspace representation to improve the performance of multi-view clustering.

\subsection{Objective Function}
Fusing the aforementioned three components jointly, the objective function of the proposed GRMSC can be written as:
\begin{equation}
\begin{array}{l}
\mathop {\min }\limits_{{\bf Z},{{\bf E}^{(k)}}} {\kern 1pt} {\kern 1pt} {\kern 1pt} {\left\| {\bf Z} \right\|_ * } + {\lambda _1}\sum\limits_{k = 1}^v {{{\left\| {{{\bf E}^{(k)}}} \right\|}_{2,1}}} \\
{\kern 1pt} {\kern 1pt} {\kern 1pt} {\kern 1pt} {\kern 1pt} {\kern 1pt} {\kern 1pt} {\kern 1pt} {\kern 1pt} {\kern 1pt} {\kern 1pt} {\kern 1pt} {\kern 1pt} {\kern 1pt} {\kern 1pt} {\kern 1pt} {\kern 1pt} {\kern 1pt} {\kern 1pt} {\kern 1pt} {\kern 1pt} {\kern 1pt} {\kern 1pt} {\kern 1pt} {\kern 1pt} {\kern 1pt} {\kern 1pt} {\kern 1pt} + {\lambda _2}\left( {{\Psi _{{\rm{ConGR}}}}({\bf Z}) + \alpha {\Psi _{{\rm{ComGR}}}}({\bf Z})} \right)\\
{\rm{s}}{\rm{.t}}{\rm{.}}{\kern 1pt} {\kern 1pt} {\kern 1pt} {{\bf X}^{(k)}} = {{\bf X}^{(k)}}{\bf Z} + {{\bf E}^{(k)}},
\end{array}
\label{Objective_Function}
\end{equation}
where $\lambda_1$, $\lambda_2$, and $\alpha$ are tradeoff parameters.

% Once the subspace representation $Z$ is attained, a similarity matrix, which is employed for spectral clustering to achieve clustering results, can be formulated as follows:
%\begin{equation}
%\frac{{{\rm{abs}}({Z^T}) + {\rm{abs}}(Z)}}{2},
%\end{equation}
%where ${\rm{abs}}( \cdot )$ indicates the element-wise absolute operator.

\section{Optimization}
To optimize the ${\bf Z}$ and ${\bf E}^{(k)}$, the ALM method \cite{ALM} is adopted and an algorithm is proposed. In order to make the optimization effectively and make the objective function separable, an auxiliary variable ${\bf Q}$ is introduced in the nuclear norm. As a result, the objective function, i.e. Eq. (\ref{Objective_Function}), can be rewritten as follows:
\begin{equation}
\begin{array}{l}
\mathop {\min }\limits_{{\bf Z},{{\bf E}^{(k)}},{\bf Q}} {\kern 1pt} {\kern 1pt} {\kern 1pt} {\left\| {\bf Q} \right\|_ * } + {\lambda _1}\sum\limits_{k = 1}^v {{{\left\| {{{\bf E}^{(k)}}} \right\|}_{2,1}}} \\
{\kern 1pt} {\kern 1pt} {\kern 1pt} {\kern 1pt} {\kern 1pt} {\kern 1pt} {\kern 1pt} {\kern 1pt} {\kern 1pt} {\kern 1pt} {\kern 1pt} {\kern 1pt} {\kern 1pt} {\kern 1pt} {\kern 1pt} {\kern 1pt} {\kern 1pt} {\kern 1pt} {\kern 1pt} {\kern 1pt} {\kern 1pt} {\kern 1pt}  + {\lambda _2}\left( {{\Psi _{{\rm{ConGR}}}}({\bf Z}) + \alpha {\Psi _{{\rm{ComGR}}}}({\bf Z})} \right)\\
{\rm{s}}{\rm{.t}}{\rm{.}}{\kern 1pt} {\kern 1pt} {\kern 1pt} {{\bf X}^{(k)}} = {{\bf X}^{(k)}}{\bf Z} + {{\bf E}^{(k)}},{\kern 1pt} {\kern 1pt} {\kern 1pt} {\bf Q} = {\bf Z},
\end{array}
\end{equation}
where ${\bf Q}$ is the auxiliary variable. And the augmented Lagrange function can be formulated:
\begin{equation}
\begin{array}{l}
{\cal L}({\bf Q},{\bf Z},{{\bf E}^{(k)}},{{\bf Y}_1^{(k)}},{\bf Y}_2)\\
{\kern 1pt} {\kern 1pt} {\kern 1pt} {\kern 1pt} {\kern 1pt} {\kern 1pt} {\kern 1pt} {\kern 1pt} {\kern 1pt} {\kern 1pt} {\kern 1pt} {\kern 1pt} {\kern 1pt} {\kern 1pt} {\kern 1pt} {\kern 1pt} {\kern 1pt} {\kern 1pt} {\kern 1pt} {\kern 1pt} {\kern 1pt} {\kern 1pt}  = {\kern 1pt} {\kern 1pt} {\kern 1pt} {\left\| {\bf Q} \right\|_ * } + {\lambda _1}\sum\limits_{k = 1}^v {{{\left\| {{{\bf E}^{(k)}}} \right\|}_{2,1}}} \\
{\kern 1pt} {\kern 1pt} {\kern 1pt} {\kern 1pt} {\kern 1pt} {\kern 1pt} {\kern 1pt} {\kern 1pt} {\kern 1pt} {\kern 1pt} {\kern 1pt} {\kern 1pt} {\kern 1pt} {\kern 1pt} {\kern 1pt} {\kern 1pt} {\kern 1pt} {\kern 1pt} {\kern 1pt} {\kern 1pt} {\kern 1pt} {\kern 1pt}  + {\lambda _2}\left( {{\Psi _{{\rm{ConGR}}}}({\bf Z}) + \alpha {\Psi _{{\rm{ComGR}}}}({\bf Z})} \right)\\
{\kern 1pt} {\kern 1pt} {\kern 1pt} {\kern 1pt} {\kern 1pt} {\kern 1pt} {\kern 1pt} {\kern 1pt} {\kern 1pt} {\kern 1pt} {\kern 1pt} {\kern 1pt} {\kern 1pt} {\kern 1pt} {\kern 1pt} {\kern 1pt} {\kern 1pt} {\kern 1pt} {\kern 1pt} {\kern 1pt} {\kern 1pt} {\kern 1pt}  + \sum\limits_{k = 1}^v {\Gamma ({\bf Y}_1^{(k)},{{\bf X}^{(k)}} - {{\bf X}^{(k)}}{\bf Z} - {{\bf E}^{(k)}})} \\
{\kern 1pt} {\kern 1pt} {\kern 1pt} {\kern 1pt} {\kern 1pt} {\kern 1pt} {\kern 1pt} {\kern 1pt} {\kern 1pt} {\kern 1pt} {\kern 1pt} {\kern 1pt} {\kern 1pt} {\kern 1pt} {\kern 1pt} {\kern 1pt} {\kern 1pt} {\kern 1pt} {\kern 1pt} {\kern 1pt} {\kern 1pt} {\kern 1pt}  + \Gamma ({{\bf Y}_2},{\bf Z} - {\bf Q}),
\end{array}
\label{ALM_Function}
\end{equation}
where ${\bf Y}_1^{(v)}$ and ${\bf Y}_2$ indicate Lagrange multipliers, and to make the representation concise, $\Gamma({\bf A},{\bf B})$ has the following definition:
\begin{equation}
\Gamma ({\bf A},{\bf B}) = \left\langle {{\bf A},{\bf B}} \right\rangle  + \frac{\mu }{2}\left\| {\bf B} \right\|_F^2
\end{equation}
where $\mu$ denotes an adaptive penalty parameter with a positive value, $\left\langle { \cdot , \cdot } \right\rangle $ is the inner product operation. Consequently, problem of minimizing the augmented Lagrange function (\ref{ALM_Function}) can be divided into four subproblems. Algorithm 1 presents the whole procedure of the optimization.

\subsection{Subproblem of Updating ${\bf E}^{(k)}$}
By fixing other variables, the subproblem with respect to ${\bf E}^{(k)}$ can be constructed:
\begin{equation}
\mathop {\min }\limits_{{{\bf E}^{(k)}}} {\kern 1pt} {\kern 1pt} {\kern 1pt} {\lambda _1}\sum\limits_{k = 1}^v {{{\left\| {{{\bf E}^{(k)}}} \right\|}_{2,1}} + \Gamma ({\bf Y}_1^{(k)},{{\bf X}^{(k)}} - {{\bf X}^{(k)}}{\bf Z} - {{\bf E}^{(k)}})}, 
\end{equation}
which can be simplified as follows:
\begin{equation}
\mathop {\min }\limits_{{{\bf E}^{(k)}}} {\kern 1pt} {\kern 1pt} {\kern 1pt} {\lambda _1}\sum\limits_{k = 1}^v {{{\left\| {{{\bf E}^{(k)}}} \right\|}_{2,1}} + \frac{\mu }{2}\left\| {{{\bf E}^{(k)}} - {\bf T}_E^{(k)}} \right\|_F^2},
\label{E_Optimization}
\end{equation}
which can be solved according to Lemma 4.1 in \cite{LRR_PAMI2012}, and ${\bf T}_E^{(k)}$ has the following definition:
\begin{equation}
{\bf T}_E^{(k)} = {{\bf X}^{(k)}} - {{\bf X}^{(k)}}{\bf Z} + \frac{{{\bf Y}_1^{(k)}}}{\mu }.
\end{equation}

\subsection{Subproblem of Updating ${\bf Q}$}
In order to update ${\bf Q}$, other variables are fixed. And following subproblem can be formulated:
\begin{equation}
\mathop {\min }\limits_{\bf Q} {\kern 1pt} {\kern 1pt} {\kern 1pt} {\left\| {\bf Q} \right\|_ * } + \Gamma ({{\bf Y}_2},{\bf Z} - {\bf Q}),
\end{equation}
optimization of which is the same with the following problem:
\begin{equation}
\mathop {\min }\limits_{\bf Q} {\kern 1pt} {\kern 1pt} {\kern 1pt} {\left\| {\bf Q} \right\|_ * } + \frac{\mu }{2}\left\| {{\bf Q} - ({\bf Z} + \frac{{{{\bf Y}_2}}}{\mu })} \right\|_F^2,
\end{equation}
which has a solution with closed form:
\begin{equation}
{\bf Q} = {\bf U}{{S}_{{1 \mathord{\left/
 {\vphantom {1 \mu }} \right.
 \kern-\nulldelimiterspace} \mu }}}({\bf\Sigma} ){\bf V},
 \label{Q_Optimization}
\end{equation}
where ${\bf U}{\bf\Sigma} {\bf V} = {\bf Z} + \frac{{{{\bf Y}_2}}}{\mu }$ and ${S_\varepsilon }$ denotes a soft-threshold operator \cite{SVT_nuclear_norm} as follows:
\begin{equation}
{S_\varepsilon }(x) = \left\{ \begin{array}{l}
x - \varepsilon ,{\kern 1pt} {\kern 1pt} {\kern 1pt} {\rm{if}}{\kern 1pt} {\kern 1pt} {\kern 1pt} x - \varepsilon  > 0\\
x + \varepsilon ,{\kern 1pt} {\kern 1pt} {\kern 1pt} {\rm{if}}{\kern 1pt} {\kern 1pt} {\kern 1pt} x - \varepsilon  < 0\\
{\kern 1pt} {\kern 1pt} {\kern 1pt} {\kern 1pt} {\kern 1pt} {\kern 1pt} {\kern 1pt} {\kern 1pt} 0{\kern 1pt} {\kern 1pt} {\kern 1pt} {\kern 1pt} {\kern 1pt} {\kern 1pt} {\kern 1pt} {\kern 1pt} {\kern 1pt} {\kern 1pt} ,{\kern 1pt} {\kern 1pt} {\kern 1pt} {\rm{otherwise}}.
\end{array} \right.
\end{equation}

\subsection{Subproblem of Updating ${\bf Z}$}
When other variables are fixed, the subproblem of Updating ${\bf Z}$ can be written as follows:
\begin{equation}
\begin{array}{l}
{\kern 1pt} \mathop {\min }\limits_{\bf Z} {\kern 1pt} {\kern 1pt} {\kern 1pt} {\lambda _2}\left( {{\Psi _{{\rm{ConGR}}}}({\bf Z}) + \alpha {\Psi _{{\rm{ComGR}}}}({\bf Z})} \right)\\
{\kern 1pt} {\kern 1pt} {\kern 1pt} {\kern 1pt} {\kern 1pt} {\kern 1pt} {\kern 1pt} {\kern 1pt} {\kern 1pt} {\kern 1pt} {\kern 1pt} {\kern 1pt} {\kern 1pt} {\kern 1pt} {\kern 1pt} {\kern 1pt} {\kern 1pt} {\kern 1pt} {\kern 1pt} {\kern 1pt} {\kern 1pt} {\kern 1pt}  + \sum\limits_{k = 1}^v {\Gamma ({\bf Y}_1^{(k)},{{\bf X}^{(k)}} - {{\bf X}^{(k)}}{\bf Z} - {{\bf E}^{(k)}})} \\
{\kern 1pt} {\kern 1pt} {\kern 1pt} {\kern 1pt} {\kern 1pt} {\kern 1pt} {\kern 1pt} {\kern 1pt} {\kern 1pt} {\kern 1pt} {\kern 1pt} {\kern 1pt} {\kern 1pt} {\kern 1pt} {\kern 1pt} {\kern 1pt} {\kern 1pt} {\kern 1pt} {\kern 1pt} {\kern 1pt} {\kern 1pt} {\kern 1pt}  + \Gamma ({{\bf Y}_2},{\bf Z} - {\bf Q}),
\end{array}
\label{Z_Subproblem}
\end{equation}
solution of which can be obtained by taking derivation with respect to ${\bf Z}$ and setting to be zeros. Specifically, to make the optimization effectively, we define a matrix ${{\bf W}}^{(k)}$:
\begin{equation}
\left\{ \begin{array}{l}
{{\bf W}}_{ij}^{(k)} = \frac{1}{v}{\bf\Lambda}_{ij}^ *,{\kern 1pt} {\kern 1pt} {\kern 1pt} {\kern 1pt} {\kern 1pt} {\kern 1pt} (i,j) \in \Omega \\
{{\bf W}}_{ij}^{(k)} = \alpha {\bf\Upsilon}_{ij}^{(k)},{\kern 1pt}{\kern 1pt} {\kern 1pt} {\kern 1pt} (i,j) \in \bar \Omega, 
\end{array} \right.
\end{equation}
and it is easy to prove the following equation:
\begin{equation}
{\Psi _{{\rm{ConGR}}}}({\bf Z}) + \alpha {\Psi _{{\rm{ComGR}}}}({\bf Z}){\rm{ = }}\sum\limits_{k = 1}^v {Tr({{\bf Z}^T}{{\bf L}^{(k)}}{\bf Z})}
\end{equation}
where ${\bf L}^{(k)}$ is the Laplacian matrix of ${{\bf W}}^{(k)}$, and ${\bf Z}^T$ indicates the transpose of the subspace representation ${\bf Z}$. Therefore, the optimization of Eq. (\ref{Z_Subproblem}) can be written as follows:
\begin{equation}
{\bf Z} = {\bf T}_{ZA}^{ - 1}{{\bf T}_{ZB}},
\label{Z_Optimization}
\end{equation}
where ${\bf T}_{ZA}^{ - 1}$ is the inverse matrix of ${\bf T}_{ZA}$, ${\bf T}_{ZA}$ and ${\bf T}_{ZB}$ have the following definition:
\begin{equation}
\begin{array}{l}
{{\bf T}_{ZA}} = {\lambda _2}\sum\limits_{k = 1}^v {\left( {{{\bf L}^{{{(k)}^T}}} + {{\bf L}^{(k)}}} \right)} \\
{\kern 1pt} {\kern 1pt} {\kern 1pt} {\kern 1pt} {\kern 1pt} {\kern 1pt} {\kern 1pt} {\kern 1pt} {\kern 1pt} {\kern 1pt} {\kern 1pt} {\kern 1pt} {\kern 1pt} {\kern 1pt} {\kern 1pt} {\kern 1pt} {\kern 1pt} {\kern 1pt} {\kern 1pt}  + \mu \left( {\sum\limits_{k = 1}^v {\left( {{{\bf X}^{{{(k)}^T}}} {{\bf X}^{(k)}}} \right)}  + {\bf I}} \right),\\
{{\bf T}_{ZB}} = \sum\limits_{k = 1}^v {\left( {{{\bf X}^{{{(k)}^T}}}{\bf Y}_1^{(k)} + \mu \left( {{{\bf X}^{{{(k)}^T}}} {{\bf X}^{(k)}}} \right)} \right)} \\
{\kern 1pt} {\kern 1pt} {\kern 1pt} {\kern 1pt} {\kern 1pt} {\kern 1pt} {\kern 1pt} {\kern 1pt} {\kern 1pt} {\kern 1pt} {\kern 1pt} {\kern 1pt} {\kern 1pt} {\kern 1pt} {\kern 1pt} {\kern 1pt} {\kern 1pt} {\kern 1pt} {\kern 1pt}  + \mu \left( {\sum\limits_{k = 1}^v {\left( {{{\bf X}^{{{(k)}^T}}}{{\bf E}^{(k)}}} \right)}  + {\bf Q}} \right),
\end{array}
\end{equation}
where ${\bf I}$ is the identity matrix with suitable size.

\subsection{Subproblem of Updating ${\bf Y}_1^{(k)}$, ${\bf Y}_2$ and $\mu$}
We update Lagrange multiplers and $\mu$ with the following form according to \cite{ALM}:
\begin{equation}
\left\{ \begin{array}{l}
{\bf Y}_1^{(k)} = {\bf Y}_1^{(k)} + \mu ({{\bf X}^{(k)}} - {{\bf X}^{(k)}}{\bf Z} - {{\bf E}^{(k)}})\\
{{\bf Y}_2} = {{\bf Y}_2} + \mu ({\bf Z} - {\bf Q})\\
\mu {\rm{ = }}\min (\rho \mu ,{\mu _{\max }}),
\end{array} \right.
\label{Y_Optimization}
\end{equation}
where $\mu _{\max }$ is a threshold value and $\rho$ indicates a nonnegative scalar.

\begin{algorithm}
\caption{Algorithm of GRMSC}
{\bf Input:} \\
\hspace*{0.12in} Multi-view $\{ {{\bf X}^{(k)}}\} _{k = 1}^v$, ${\bf E}^{(k)}={\bf 0}$, ${\bf Q}={\bf 0}$\\
\hspace*{0.12in} ${\bf Y}_1^{(k)}={\bf 0}$, ${\bf Y}_2={\bf 0}$, ${\bf Z}$ with random initialization,\\
\hspace*{0.12in} $\rho=1.9$, $\mu  = {10^{ - 4}}$, ${\mu _{\max }} = {10^{6}}$, $\varepsilon  = {10^{ - 6}}$;\\
{\bf Output:} \\ 
\hspace*{0.12in} ${\bf Z}$;\\
{\bf Repeat:}\\
\hspace*{0.12in} {\bf For} $k = 1,2 \cdots ,v$ {\bf do}:\\
\hspace*{0.36in} Updating ${\bf E}^{(k)}$ according to (\ref{E_Optimization});\\
\hspace*{0.12in} {\bf End}\\
\hspace*{0.12in} Updating ${\bf Q}$ according to (\ref{Q_Optimization});\\
\hspace*{0.12in} Updating ${\bf Z}$ according to (\ref{Z_Optimization});\\
\hspace*{0.12in} {\bf For} $k = 1,2 \cdots ,v$ {\bf do}:\\
\hspace*{0.36in} Updating ${\bf Y}_1^{(k)}$ according to (\ref{Y_Optimization});\\
\hspace*{0.12in} {\bf End}\\
\hspace*{0.12in} Updating ${\bf Y}_2$ and $\mu$ according to (\ref{Y_Optimization});\\
{\bf Until:}\\
\hspace*{0.12in} {\bf For} $k = 1,2 \cdots ,v$:\\
\hspace*{0.36in} ${\left\| {{{\bf X}^{(k)}} - {{\bf X}^{(k)}}{{\bf Z}} - {\bf E}^{(k)}} \right\|_\infty } < \varepsilon, $\\
\hspace*{0.12in} {\bf End}\\
\hspace*{0.12in} and ${\left\| {{\bf Z} - {\bf Q}} \right\|_\infty } < \varepsilon. $
\end{algorithm}

\subsection{Computational Complexity}
% Convergence property of our algorithm is analyzed firstly. Since the objective function is decomposed into more than two subproblems, it is hard to give a solid proof of the convergence \cite{ALM}. Fortunately, experimental results in next section demonstrate the good convergence of the proposed algorithm.

The main computational burden is consist of the four subproblems. Besides, ${\bf\Lambda}^{(k)}$, ${\bf\Lambda}^ *$ and ${\bf\Upsilon}^{(k)}$ are pre-computed outside of the algorithm. In line with Table \ref{table_symbol}, the number of samples is $n$, the number of views is $v$, the number of iteration is $t$, and the dimension of the $k$th view is $d_k$. For convenience, $d$ is introduced and $d = \max (\left\{ {{d_k}} \right\}_{k = 1}^v)$. The complexity of updating $\left\{ {{{\bf E}^{(k)}}} \right\}_{k = 1}^v$ and ${\bf Q}$ are ${\cal O}(vdn)$ and ${\cal O}(n^3)$ respectively, as for updating ${\bf Z}$ and Lagrange multiplers, the complexity is ${\cal O}(n^3+vdn)$. Therefore, the computational complexity of Algorithm 1 is ${\cal O}(tn(vd+n^2))$.

\begin{table*}[t]
\centering
\caption{Clustering results of the validation and ablation experiments.}\smallskip
\resizebox{1\textwidth}{!}{
\begin{tabular}{l|l|l|l|l|l|l|l}
\hline
Dataset & Method & NMI & ACC & F-Score & AVG & Precious & RI \\ 
\hline
\multirow{4}{*}{3-Sources}
& LRR$\rm{_{BSV}}$ & 0.6348(0.0078) & 0.6783(0.0136) & 0.6158(0.0185) & 0.7958(0.0150) & 0.6736(0.0148) & 0.8356(0.0070)\\
& MSC$\rm{_{Naive}}$ & 0.6307(0.0075) & 0.7079(0.0098) & 0.6526(0.0101) & 0.8167(0.0233) & 0.6959(0.0130) & 0.8479(0.0044)\\
& GRMSC$\rm{_{Naive}}$ & 0.6726(0.0099) & 0.7012(0.0111) & 0.6447(0.0089) & 0.6859(0.0243) & \bf 0.7335(0.0090) & 0.8527(0.0035)\\
& GRMSC & \bf 0.7321(0.0068) & \bf 0.7799(0.0025) & \bf 0.7359(0.0036) & \bf 0.6163(0.0173) & 0.7288(0.0057) & \bf 0.8760(0.0021)\\
\hline
\multirow{4}{*}{BBCSport}
& LRR$\rm{_{BSV}}$ & 0.6996(0.0000) & 0.7970(0.0015) & 0.7612(0.0001) & 0.7269(0.0006) & 0.6890(0.0001) & 0.8727(0.0000)\\
& MSC$\rm{_{Naive}}$ & 0.8379(0.0000) & 0.9099(0.0000) & 0.8968(0.0000) & 0.3661(0.0000) & 0.8914(0.0000) & 0.9505(0.0000)\\
& GRMSC$\rm{_{Naive}}$ & 0.8425(0.0000) & 0.9118(0.0000) & 0.9011(0.0000) & 0.3567(0.0000) & 0.8948(0.0000) & 0.9525(0.0000)\\
& GRMSC & \bf 0.8985(0.0000) & \bf 0.9669(0.0000) & \bf 0.9330(0.0000) & \bf 0.2152(0.0000) & \bf 0.9418(0.0000) & \bf 0.9683(0.0000)\\
\hline
\multirow{4}{*}{Movie 617}
& LRR$\rm{_{BSV}}$ & 0.2690(0.0063) & 0.2767(0.0093) & 0.1566(0.0040) & 2.9462(0.0250) & 0.1528(0.0042) & 0.8943(0.0015)\\
& MSC$\rm{_{Naive}}$ & 0.2765(0.0043) & 0.2644(0.0040) & 0.1544(0.0022) & 2.9159(0.0173) & 0.1519(0.0024) & 0.8949(0.0009)\\
& GRMSC$\rm{_{Naive}}$ & 0.3344(0.0065) & 0.3159(0.0087) & 0.2114(0.0075) & 2.6897(0.0264) & 0.2020(0.0082) & 0.8989(0.0023)\\
& GRMSC & \bf 0.3367(0.0084) & \bf 0.3209(0.0128) & \bf 0.2135(0.0132) & \bf 2.6816(0.0319) & \bf 0.2040(0.0104) & \bf 0.8992(0.0023)\\
\hline
\multirow{4}{*}{NGs}
& LRR$\rm{_{BSV}}$ & 0.3402(0.0201) & 0.4213(0.0184) & 0.3911(0.0056) & 1.7056(0.0461) & 0.2688(0.0065) & 0.5556(0.0199)\\
& MSC$\rm{_{Naive}}$ & 0.9096(0.0000) & 0.9700(0.0000) & 0.9410(0.0000) & 0.2101(0.0000) & 0.9408(0.0000) & 0.9766(0.0000)\\
& GRMSC$\rm{_{Naive}}$ & 0.9217(0.0000) & 0.9740(0.0000) & 0.9488(0.0000) & 0.1819(0.0000) & 0.9485(0.0000) & 0.9797(0.0000)\\
& GRMSC & \bf 0.9547(0.0000) & \bf 0.9860(0.0000) & \bf 0.9721(0.0000) & \bf 0.1052(0.0000) & \bf 0.9720(0.0000) & \bf 0.9889(0.0000)\\
\hline
\multirow{4}{*}{Prokaryotic}
& LRR$\rm{_{BSV}}$ & 0.4462(0.0000) & \bf 0.7822(0.0000) & \bf 0.7167(0.0000) & 0.8946(0.0000) & 0.7207(0.0000) & 0.7779(0.0000)\\
& MSC$\rm{_{Naive}}$ & 0.3602(0.0004) & 0.6915(0.0000) & 0.6015(0.0001) & 1.0401(0.0007) & 0.5797(0.0001) & 0.6735(0.0001)\\
& GRMSC$\rm{_{Naive}}$ & 0.4187(0.0000) & 0.5989(0.0000) & 0.5459(0.0000) & 0.9018(0.0000) & 0.6083(0.0000) & 0.6753(0.0000)\\
& GRMSC & \bf 0.5054(0.0000) & 0.7731(0.0000) & 0.6922(0.0000) & \bf 0.7372(0.0000) & \bf 0.7936(0.0000) & \bf 0.7848(0.0000)\\
\hline
\multirow{4}{*}{Yale Face}
& LRR$\rm{_{BSV}}$ & 0.7134(0.0098) & 0.7034(0.0125) & 0.5561(0.0159) & 1.1328(0.0390) & 0.5404(0.0176) & 0.9442(0.0022)\\
& MSC$\rm{_{Naive}}$ & 0.6877(0.0109) & 0.6352(0.0215) & 0.4819(0.0166) & 1.2545(0.0433) & 0.4487(0.0175) & 0.9317(0.0026)\\
& GRMSC$\rm{_{Naive}}$ & 0.7680(0.0342) & 0.7362(0.0467) & 0.6282(0.0458) & 0.9224(0.1338) & 0.6092(0.0466) & 0.9531(0.0060)\\
& GRMSC & \bf 0.7709(0.0306) & \bf 0.7418(0.0339) & \bf 0.6306(0.0425) & \bf 0.9107(0.1198) & \bf 0.6115(0.0437) & \bf 0.9535(0.0056)\\
\hline
\end{tabular}
}
\label{Validataion_Expreimental_Results}
\end{table*}

\section{Experiments}
Comprehensive experiments are conducted and presented in this section. Furthermore, the convergence property and parameter sensitivity of the proposed method are analyzed as well. Six benchmark datasets are employed. In particular, 3-Sources \cite{GMC} is a three-view dataset containing news article data from BBC, Reuters, and Guardian. BBCSport \cite{RMSC} consists of 544 sports news reports, each of which is decomposed into two subparts. Movie617 contains 617 movie samples of 17 categories with two views, i.e., keywords and actors. NGs \cite{GMC} consisting of 500 samples is a subset of the 20 Newsgroup datasets and has three views. Prokaryotic \cite{MLRSSC_PR2018} is a multi-view dataset that describes prokaryotic species from three aspects: textual data, proteome composition, and genomic representations. Yale Face is a dataset containing 165 face images of 15 individuals and each image is described by three features, namely intensity, LBP, and Gabor. Additionally, six evaluation metrics \cite{metric_reference,RMSC,LMSC} are utilized: Normalized Mutual Information (NMI), ACCuracy (ACC), F-Score, AVGent (AVG), Precious, and Rand Index (RI). Higher values of all metrics, except for AVGent, demonstrate the better clustering results. Parameters of all comparison methods are fine-tuned. To eliminate the randomness, 30 test runs with random initialization are performed and clustering results are represented in the form of mean values with standard derivation. The numbers in the bold type denote the best clustering results.

\subsection{Validation and Ablation Experiments}
To validate the effectiveness of our GRMSC, results of three different methods are compared. The first clustering method is based on the low-rank representation \cite{LRR_PAMI2012} with best single view, i.e., LRR$\rm{_{BSV}}$. The second clustering method is based on the subspace representation obtained from Eq. (\ref{LRR_MSC_Only}), named MSC$\rm{_{Naive}}$ for convenience. The third method is the graph-regularized multi-view subspace clustering, which only leverages the first-order proximity to construct the graph regularizer and is termed the GRMSC$\rm{_{Naive}}$.

As displayed in Table \ref{Validataion_Expreimental_Results}, multi-view clustering can generally achieve better clustering results than those of single view clustering. Furthermore, compared with MSC$\rm{_{Naive}}$ and GRMSC$\rm{_{Naive}}$, the proposed GRMSC method achieves significantly better clustering performance, which validates the necessity of introducing the consistent graph regularizer and the complementary graph regularizer, while verifying the effectiveness of the proposed method.

\begin{figure}[t]
\centering
\includegraphics[width=0.75\columnwidth]{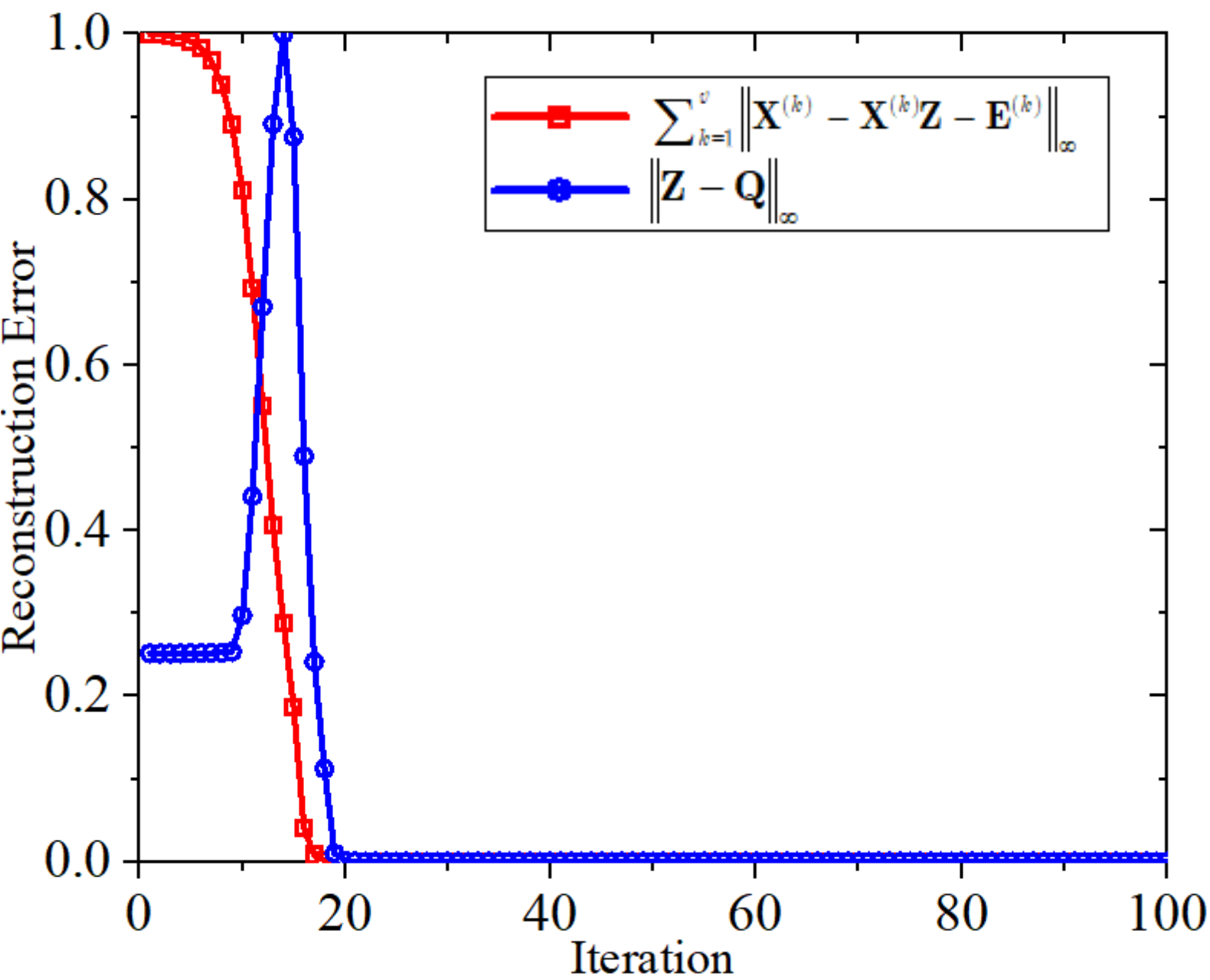} % Reduce the figure size so that it is slightly narrower than the column. Don't use precise values for figure width.This setup will avoid overfull boxes. 
\caption{Convergence property of GRMSC on NGs, Obverously, two curves attain convergence within 20 iterations.}
\label{Convergence}
\end{figure}

\begin{figure}[t]
\centering
\includegraphics[width=0.75\columnwidth]{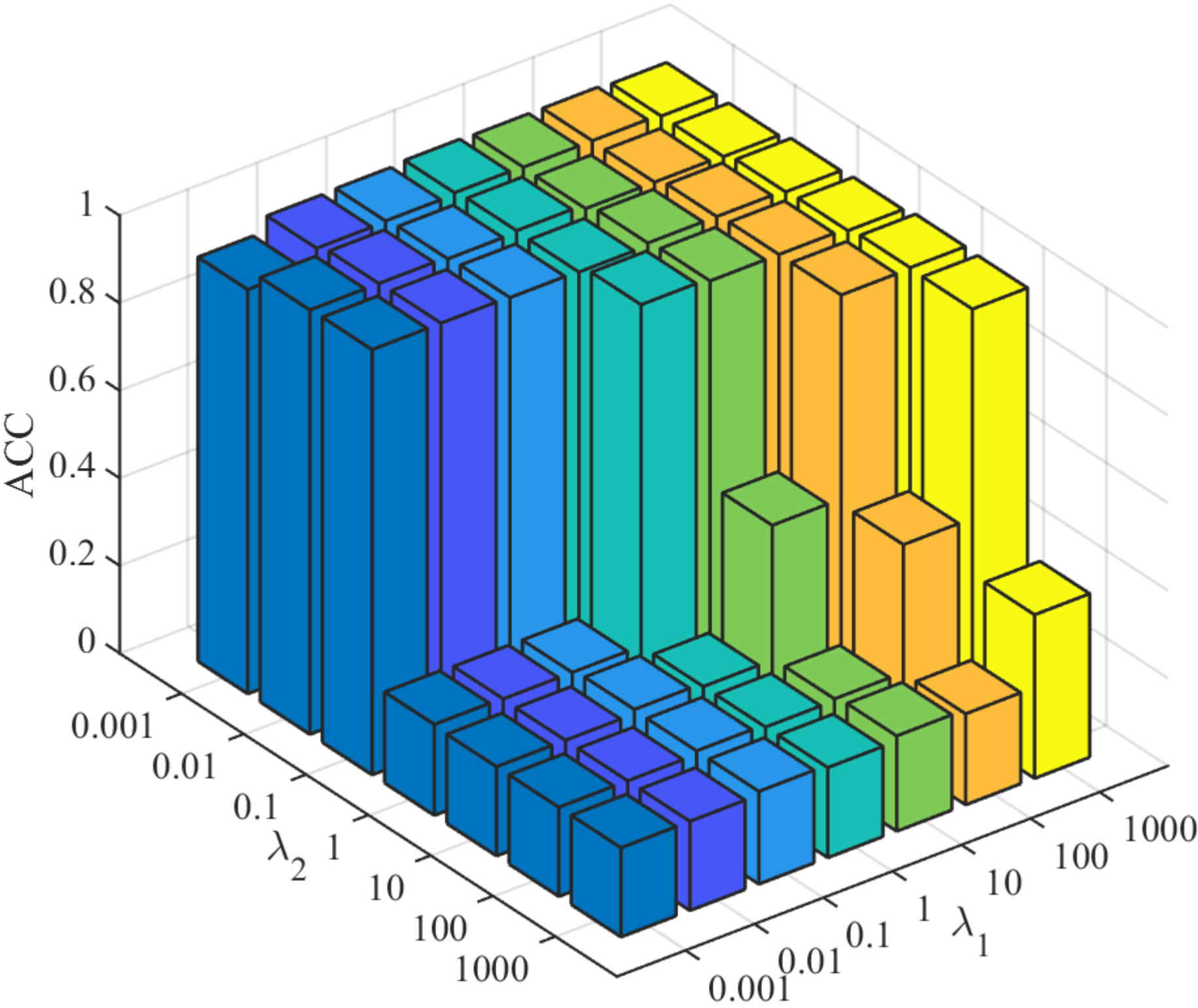} % Reduce the figure size so that it is slightly narrower than the column. Don't use precise values for figure width.This setup will avoid overfull boxes. 
\caption{Clustering results in metric of NMI with respect to different $\lambda_1$ and $\lambda_2$ on NGs.}
\label{Parameter_Sensitive}
\end{figure}

\begin{table*}[t]
\centering
\caption{Clustering results of the comparison experiments.}\smallskip
\resizebox{1\textwidth}{!}{
\begin{tabular}{l|l|l|l|l|l|l|l}
\hline
Dataset & Method & NMI & ACC & F-Score & AVG & Precious & RI \\ 
\hline
\multirow{6}{*}{3-Sources}
& RMSC \cite{RMSC} & 0.5109(0.0100) & 0.5379(0.0108) & 0.4669(0.0097) & 1.0946(0.0254) & 0.4970(0.0136) & 0.7650(0.0054)\\
& AMGL \cite{AMGL} & 0.5865(0.0510) & 0.6726(0.0394) & 0.5895(0.0414) & 1.0841(0.1344) & 0.4865(0.0592) & 0.7517(0.0496)\\
& LMSC \cite{LMSC} & 0.6748(0.0195) & 0.7059(0.0198) & 0.6451(0.0177) & 0.6827(0.0496) & \bf 0.7314(0.0237) & 0.8524(0.0081)\\
& MLRSSC \cite{MLRSSC_PR2018} & 0.5919(0.0025) & 0.6686(0.0000) & 0.6353(0.0011) & 0.9378(0.0070) & 0.6410(0.0018) & 0.8320(0.0007)\\
& GMC \cite{GMC} & 0.6216(0.0000) & 0.6923(0.0000) & 0.6047(0.0000) & 1.0375(0.0000) & 0.4844(0.0000) & 0.7556(0.0000)\\
& GRMSC & \bf 0.7321(0.0068) & \bf 0.7799(0.0025) & \bf 0.7359(0.0036) & \bf 0.6163(0.0173) & 0.7288(0.0057) & \bf 0.8760(0.0021)\\
\hline
\multirow{6}{*}{BBCSport}
& RMSC \cite{RMSC} & 0.8124(0.0074) & 0.8562(0.0198) & 0.8514(0.0132) & 0.4159(0.0149) & 0.8566(0.0105) & 0.9297(0.0065)\\
& AMGL \cite{AMGL} & 0.8640(0.0681) & 0.9189(0.0870) & 0.9008(0.0868) & 0.3305(0.1858) & 0.8708(0.1188) & 0.9477(0.0513)\\
& LMSC \cite{LMSC} & 0.8393(0.0043) & 0.9180(0.0031) & 0.8996(0.0033) & 0.3608(0.0094) & 0.8938(0.0036) & 0.9518(0.0016)\\
& MLRSSC \cite{MLRSSC_PR2018} & 0.8855(0.0000) & 0.9651(0.0000) & 0.9296(0.0000) & 0.2437(0.0000) & 0.9384(0.0000) & 0.9667(0.0000)\\
& GMC \cite{GMC} & 0.7954(0.0000) & 0.7390(0.0000) & 0.7207(0.0000) & 0.6450(0.0000) & 0.5728(0.0000) & 0.8204(0.0000)\\
& GRMSC & \bf 0.8985(0.0000) & \bf 0.9669(0.0000) & \bf 0.9330(0.0000) & \bf 0.2152(0.0000) & \bf 0.9418(0.0000) & \bf 0.9683(0.0000)\\
\hline
\multirow{6}{*}{Movie 617}
& RMSC \cite{RMSC} & 0.2969(0.0023) & 0.2986(0.0043) & 0.1819(0.0024) & 2.8498(0.0095) & 0.1674(0.0024) & 0.8903(0.0012)\\
& AMGL \cite{AMGL} & 0.2606(0.0088) & 0.2563(0.0124) & 0.1461(0.0055) & 3.1105(0.0387) & 0.0971(0.0063) & 0.7845(0.0272)\\
& LMSC \cite{LMSC} & 0.2796(0.0096) & 0.2694(0.0133) & 0.1601(0.0088) & 2.9129(0.0388) & 0.1512(0.0092) & 0.8909(0.0030)\\
& MLRSSC \cite{MLRSSC_PR2018} & 0.2975(0.0061) & 0.2887(0.0111) & 0.1766(0.0068) & 2.8481(0.0216) & 0.1619(0.0064) & 0.8893(0.0023)\\
& GMC \cite{GMC} & 0.2334(0.0000) & 0.1864(0.0000) & 0.1242(0.0000) & 3.3795(0.0000) & 0.0682(0.0000) & 0.3995(0.0000)\\
& GRMSC & \bf 0.3367(0.0084) & \bf 0.3209(0.0128) & \bf 0.2135(0.0132) & \bf 2.6816(0.0319) & \bf 0.2040(0.0104) & \bf 0.8992(0.0023)\\
\hline
\multirow{6}{*}{NGs}
& RMSC \cite{RMSC} & 0.1580(0.0099) & 0.3700(0.0081) & 0.3070(0.0058) & 1.9755(0.0236) & 0.2664(0.0074) & 0.6726(0.0086)\\
& AMGL \cite{AMGL} & 0.8987(0.0464) & 0.9393(0.0903) & 0.9212(0.0709) & 0.2473(0.1385) & 0.9088(0.1024) & 0.9665(0.0339)\\
& LMSC \cite{LMSC} & 0.9052(0.0075) & 0.9705(0.0026) & 0.9417(0.0050) & 0.2203(0.0173) & 0.9415(0.0051) & 0.9769(0.0020)\\
& MLRSSC \cite{MLRSSC_PR2018} & 0.8860(0.0000) & 0.9620(0.0000) & 0.9255(0.0000) & 0.2651(0.0000) & 0.9252(0.0000) & 0.9704(0.0000)\\
& GMC \cite{GMC} & 0.9392(0.0000) & 0.9820(0.0000) & 0.9643(0.0000) & 0.1413(0.0000) & 0.9642(0.0000) & 0.9858(0.0000)\\
& GRMSC & \bf 0.9547(0.0000) & \bf 0.9860(0.0000) & \bf 0.9721(0.0000) & \bf 0.1052(0.0000) & \bf 0.9720(0.0000) & \bf 0.9889(0.0000)\\
\hline
\multirow{6}{*}{Prokaryotic}
& RMSC \cite{RMSC} & 0.3064(0.0107) & 0.5090(0.0071) & 0.4438(0.0066) & 1.0626(0.0190) & 0.5627(0.0081) & 0.6380(0.0042)\\
& AMGL \cite{AMGL} & 0.1162(0.0522) & 0.5192(0.0195) & 0.5028(0.0164) & 1.4611(0.0826) & 0.4038(0.0328) & 0.4673(0.0652)\\
& LMSC \cite{LMSC} & 0.1485(0.0184) & 0.4233(0.0243) & 0.3663(0.0113) & 1.3520(0.0319) & 0.4397(0.0133) & 0.5718(0.0079)\\
& MLRSSC \cite{MLRSSC_PR2018} & 0.3230(0.0006) & 0.6587(0.0006) & 0.5865(0.0005) & 1.0837(0.0012) & 0.6222(0.0006) & 0.6917(0.0002)\\
& GMC \cite{GMC} & 0.1934(0.0000) & 0.4955(0.0000) & 0.4607(0.0000) & 1.3169(0.0000) & 0.4467(0.0000) & 0.5611(0.0000)\\
& GRMSC & \bf 0.5054(0.0000) & \bf 0.7731(0.0000) & \bf 0.6922(0.0000) & \bf 0.7372(0.0000) & \bf 0.7936(0.0000) & \bf 0.7848(0.0000)\\
\hline
\multirow{6}{*}{Yale Face}
& RMSC \cite{RMSC} & 0.6812(0.0089) & 0.6283(0.0146) & 0.5059(0.0119) & 1.2692(0.0365) & 0.4819(0.0137) & 0.9364(0.0019)\\
& AMGL \cite{AMGL} & 0.6437(0.0192) & 0.6046(0.0399) & 0.3986(0.0323) & 1.4710(0.0919) & 0.3378(0.0431) & 0.9087(0.0130)\\
& LMSC \cite{LMSC} & 0.7011(0.0096) & 0.6691(0.0095) & 0.5031(0.0151) & 1.2062(0.0391) & 0.4638(0.0175) & 0.9337(0.0026)\\
& MLRSSC \cite{MLRSSC_PR2018} & 0.7005(0.0311) & 0.6733(0.0384) & 0.5399(0.0377) & 1.1847(0.1206) & 0.5230(0.0378) & 0.9420(0.0049)\\
& GMC \cite{GMC} & 0.6892(0.0000) & 0.6545(0.0000) & 0.4801(0.0000) & 1.2753(0.0000) & 0.4188(0.0000) & 0.9257(0.0000)\\
& GRMSC & \bf 0.7709(0.0306) & \bf 0.7418(0.0339) & \bf 0.6306(0.0425) & \bf 0.9107(0.1198) & \bf 0.6115(0.0437) & \bf 0.9535(0.0056)\\
\hline
\end{tabular}
}
\label{Comparision_Expreimental_Results}
\end{table*}

\subsection{Comparison Experiments}
To demonstrate the superiority of the GRMSC method, Table \ref{Comparision_Expreimental_Results} displays the comparison of experimental results of five state-of-the-art multi-view subspace clustering methods, namely RMSC \cite{RMSC}, AMGL \cite{AMGL}, LMSC \cite{LMSC}, MLRSSC \cite{MLRSSC_PR2018}, GMC \cite{GMC}, previously discussed in the section \textbf{Related Works}.

The GRMSC method outperforms other methods on all benchmark datasets. For example, considering the experimental results on the Yale Face dataset, this method improves clustering performance over the second best one by approximately $6.98\%$ and $7.27\%$ with respect to NMI and ACC, respectively. It is noteworthy that although the clustering result of LMSC for the Precious metric is slightly better, GRMSC scores over the second best one by a significant margin in the remaining five metrics. Table \ref{Comparision_Expreimental_Results} displays the competitiveness of the proposed method with respect to other state-of-the-art clustering methods.

\subsection{Convergence and Parameter Sensitivity}
We consider the experiments on NGs. As depicted in Figure \ref{Convergence}, the proposed method has a stable convergence and can converge within 20 iterations. Actually, for experiments on all datasets, the proposed method has similar convergence.

Three parameters, namely $\lambda_1$, $\lambda_2$, and $\alpha$, are involved in our GRMSC. For convenience, $\alpha$, which is the parameter to balance the $\Psi _{{\rm{ConGR}}}$ and $\Psi _{{\rm{ComGR}}}$, is fixed and set as $0.001$ in this study for all datasets.  $\lambda_1$ tunes bases on the prior multi-view data information, including corruption and noise level. $\lambda_2$ is tuned to balance the importance between the low-rank representation of all views and the two graph regularizers. Furthermore, values of $\lambda_1$ and $\lambda_2$ are selected from the set $\{ 0.001,0.01,0.1,1,10,100,1000\}$. As shown in Figure \ref{Parameter_Sensitive}, good clustering results can be obtained with ${\lambda _1} \ge 1$ and ${\lambda _2} \le 1$.

\section{Conclusion}
This paper proposes a consistent and complementary graph-regularized multi-view subspace clustering to accurately integrate information from multiple views for clustering. By introducing the consistent graph regularizer and the complementary graph-regularizer, graph information of multi-view data is considered. Both the consensus and complementary information of multi-view data are fully considered for clustering. An elaborate optimization algorithm is also developed to achieve improved clustering results, and extensive experiments are conducted on six benchmark datasets to illustrate the effectiveness and competitiveness of the proposed GRMSC method in comparison to several state-of-the-art multi-view clustering methods.

\bibliographystyle{aaai}
\bibliography{AAAI2020_Ref}
% \fontsize{9.8pt}{10.8pt} \selectfont
\end{document}